\documentclass[lettersize,journal]{IEEEtran}

\usepackage{cite}
\usepackage{amsmath,amssymb,amsfonts}
\usepackage{algorithmic}
\usepackage{graphicx}
\usepackage{makecell}
\usepackage{multirow}
\usepackage{color}
\usepackage{algorithm,algorithmic}

\usepackage{textcomp}
\usepackage{wrapfig}
\usepackage[colorlinks,
            linkcolor=red,
            anchorcolor=blue,
            citecolor=green 
            ]{hyperref}

\begin{document}

\title{NiteDR: Nighttime Image De-Raining with Cross-View Sensor Cooperative Learning for Dynamic Driving Scenes}

\author{Cidan Shi, Lihuang Fang, Han Wu, Xiaoyu Xian, Yukai Shi$^\dagger$, Liang Lin,~\IEEEmembership{Fellow, IEEE}}
\maketitle

\begin{abstract}
In real-world environments, outdoor imaging systems are often affected by disturbances such as rain degradation. Especially, in nighttime driving scenes, insufficient and uneven lighting shrouds the scenes in darkness, resulting degradation of both the image quality and visibility. Particularly, in the field of autonomous driving, the visual perception ability of RGB sensors experiences a sharp decline in such harsh scenarios. Additionally, driving assistance systems suffer from reduced capabilities in capturing and discerning the surrounding environment, posing a threat to driving safety. Single-view information captured by single-modal sensors cannot comprehensively depict the entire scene. To address these challenges, we developed an image de-raining framework tailored for rainy nighttime driving scenes. It aims to remove rain artifacts, enrich scene representation, and restore useful information. Specifically, we introduce cooperative learning between visible and infrared images captured by different sensors. By cross-view fusion of these multi-source data, the scene within the images gains richer texture details and enhanced contrast. We constructed an information cleaning module called CleanNet as the first stage of our framework. Moreover, we designed an information fusion module called FusionNet as the second stage to fuse the clean visible images with infrared images. Using this stage-by-stage learning strategy, we obtain de-rained fusion images with higher quality and better visual perception. Extensive experiments demonstrate the effectiveness of our proposed Cross-View Cooperative Learning (CVCL) in adverse driving scenarios in low-light rainy environments. The proposed approach addresses the gap in the utilization of existing rain removal algorithms in specific low-light conditions. It also holds promise for extending the application of image de-raining and image fusion methods to computer vision tasks. The code is available at: \url{https://github.com/CidanShi/NiteDR-Nighttime-Image-De-raining}.
\end{abstract}


\section{Introduction}
\IEEEPARstart{R}{eal}-world interferences such as rainfall degradation often cause blurry effects, poor contrast, and color distortion, etc. During nighttime, the image quality and visibility degrade significantly due to insufficient lighting conditions. In such low-light environments, raindrops further degrades image contrast and visibility, leading to the loss of essential textures and details. This, in turn, yields suboptimal visual effects, significantly impacting the performance of high-level vision applications. Especially, in autonomous driving, rain and poor illumination hinder the acquisition of crucial environmental information, posing challenges to ensuring safe driving in rainy nighttime environments.
In computer vision tasks, image de-raining technology has attracted widespread research attention. Although recent deep learning-based algorithms~\cite{wang2022uformer},~\cite{zamir2022restormer},~\cite{frants2023qsam}, \cite{chen2023hybrid} have achieved remarkable performance, most of these methods focus on daytime scenes without taking lighting conditions into account. Faced with complex and challenging nighttime driving scenes, existing de-raining methods struggle to achieve satisfactory results. The problem of nighttime image de-raining, especially for real-world driving scenarios, remains to be solved.

Driving assistance relies on sensors for perception and navigation. In the case of heavy rain at night, the perception of RGB sensors drops sharply, resulting in severe degradation in image clarity and visibility. Inadequate lighting leads to essential details loss, preventing a comprehensive scene representation.
Although some techniques have been introduced into outdoor imaging and photography systems to improve illumination, captured images still suffer from problems such as color distortion and low contrast. Images from solitary-modal sensors offer limited scene information and emphasize specific viewpoints, while multi-modal data is complementary, providing a more intricate and comprehensive environmental perception.
Effectively leveraging these complementary data can provide more information to understand complex and challenging rainy nighttime driving scenarios. Cross-view sensor cooperation holds the potential to promote the development of nighttime de-raining algorithms, thus improving the feasibility and safety of driving on poor illumination and rainy scenes.

\begin{figure*}[!t]
\centering
\includegraphics[width=0.95\linewidth]{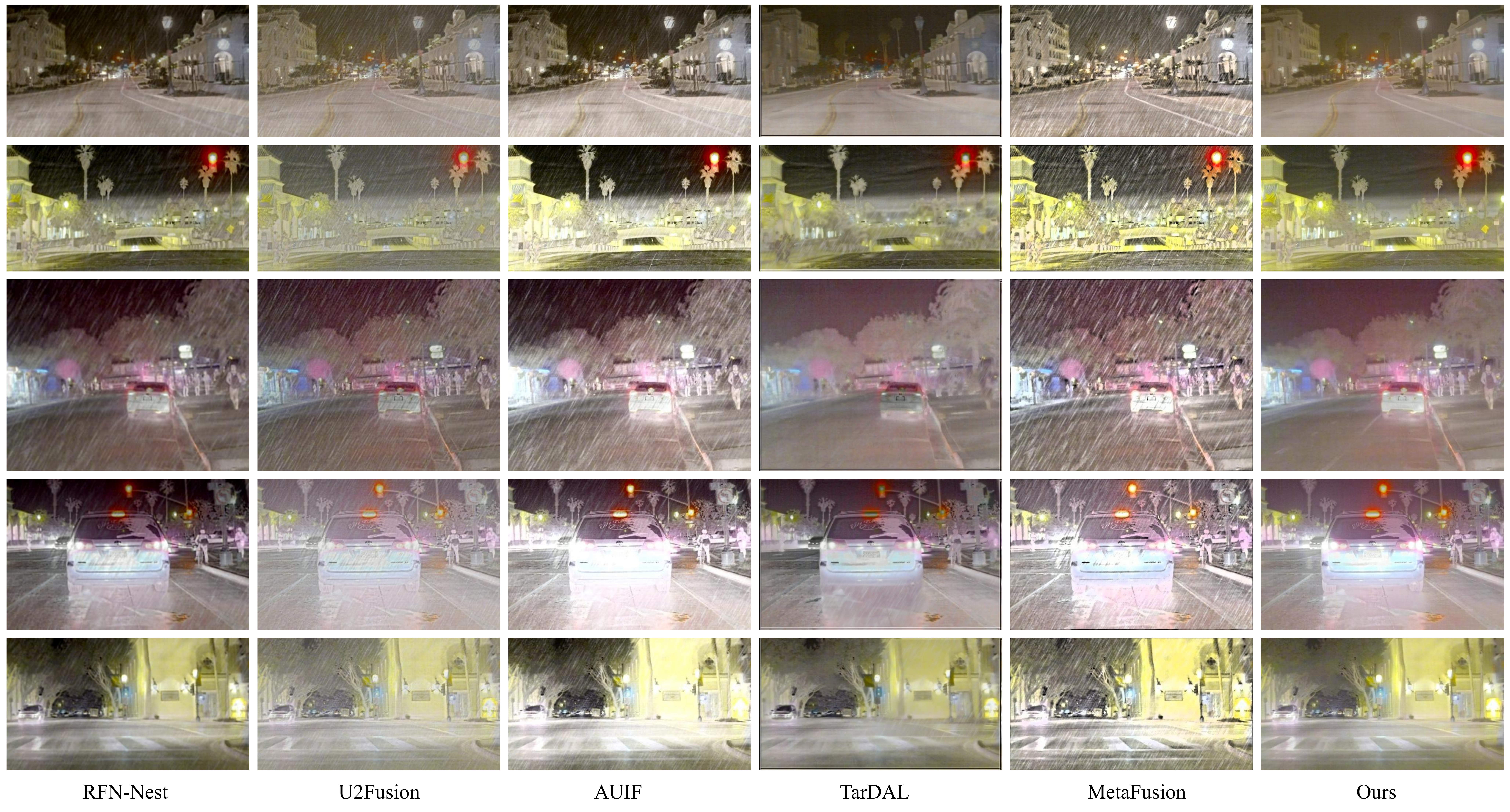}
\caption{Visual comparison of five fusion methods in rainy conditions. \textbf{As all fusion methods are easily violated by rainfall degradation, it is necessary to investigate an effective nighttime de-raining method for real-world applications.} Best viewed by zooming in the electronic version.}
\label{fig1}
\vspace{-3mm}
\end{figure*}

\begin{figure*}[t]
\centering
\includegraphics[width=0.98\linewidth]{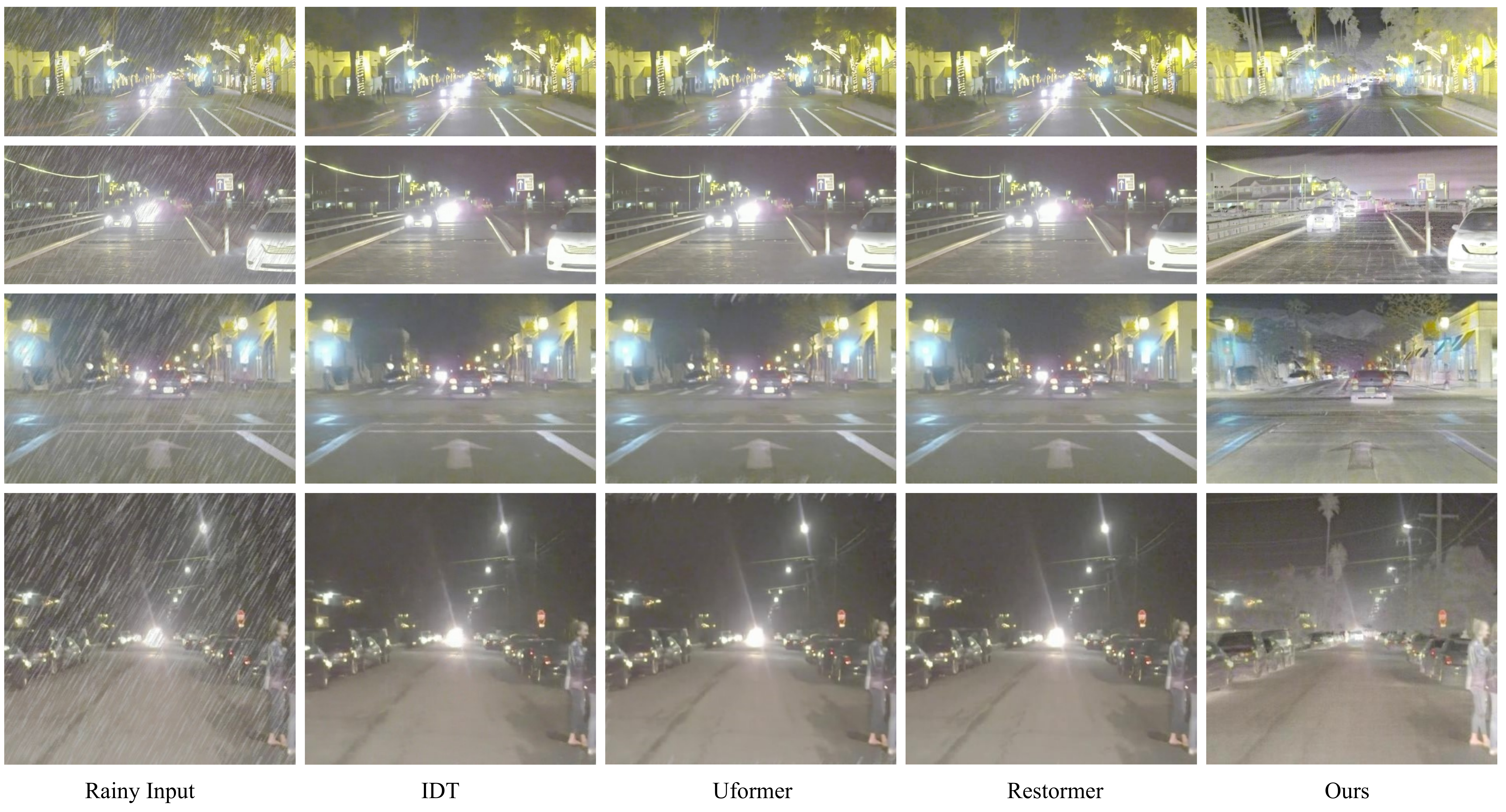}
\caption{Performance of existing de-raining methods in challenging rainy and low-light conditions. Image de-raining methods typically adopt a single sensor for perception. However, cross-sensor data has the potential to realise superior image de-raining in challenging conditions. To this end, image de-raining with multi-modal sensor data is essential to advance safe driver assistant systems. Best viewed by zooming in the electronic version. }
\label{fig2}
\vspace{-3mm}
\end{figure*}

\begin{figure}[t]
\centering
\includegraphics[width=0.95\columnwidth]{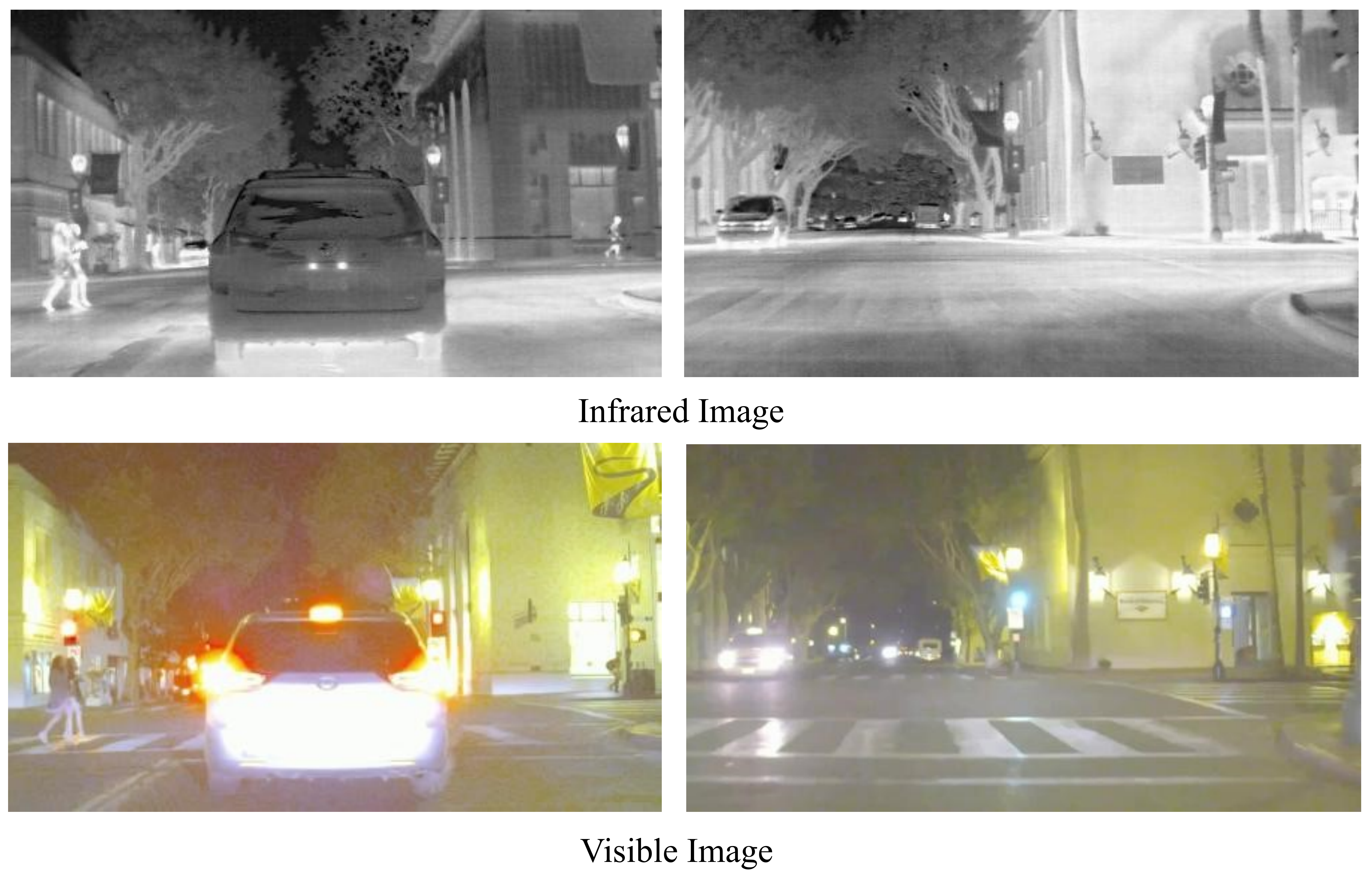}
\caption{Formerly, image fusion methods assume all data is clear and ignore the rainfall degradation in the real-world condition.}
\label{fig3}
\vspace{-3mm}
\end{figure}

\begin{figure}[t]
\centering
\includegraphics[width=0.99\columnwidth]{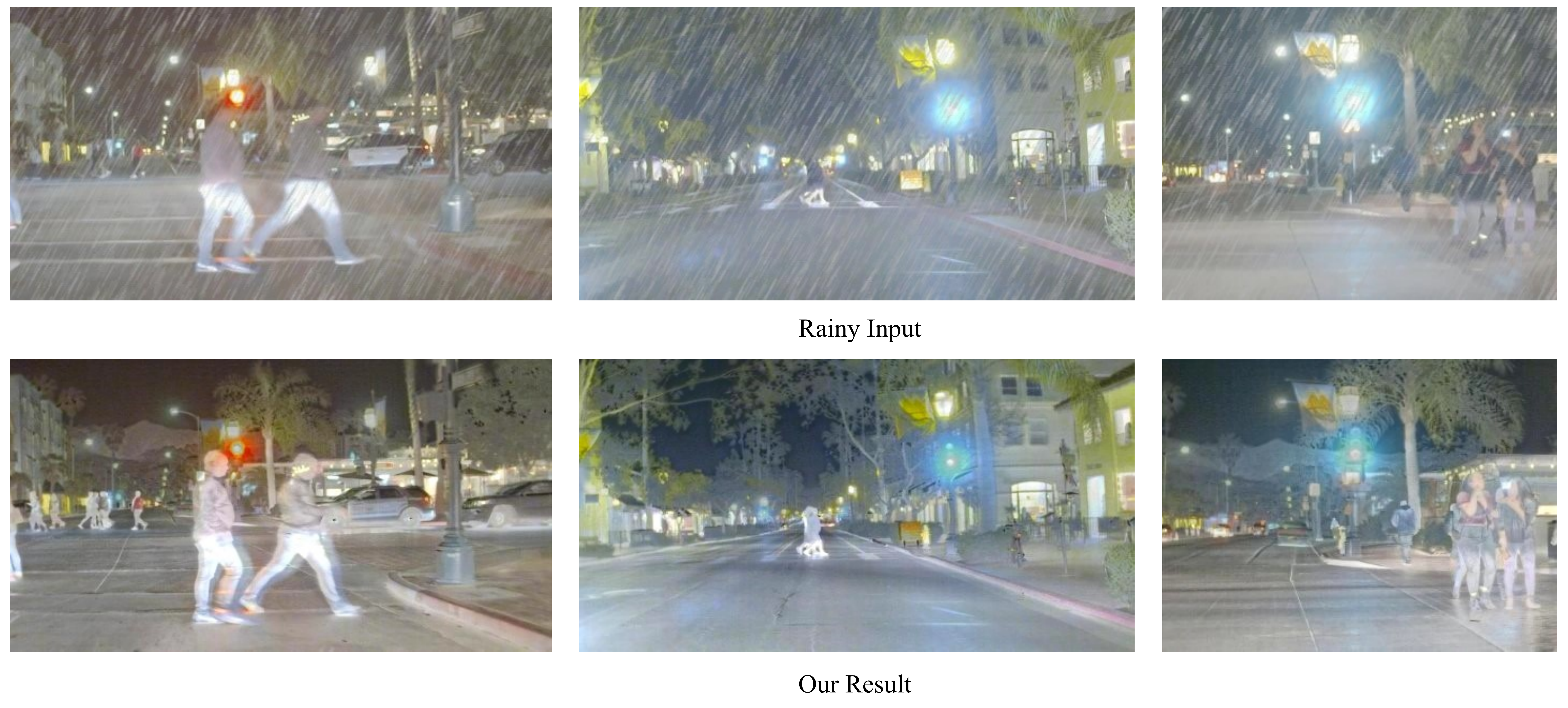}
\caption{With cross-view sensor fusion network, our algorithm efficiently removes rainwater, revealing detailed textures and prominent targets in low-light conditions. Best viewed by zooming in the electronic version.}
\label{fig4}
\vspace{-3mm}
\end{figure}

Observing visible and infrared data, we aim to leverage both for a suitable de-raining framework for nighttime driving scenes through cross-view sensor cooperative learning. Cross-sensor information fusion can preserve the merits of each modality, generating richer fusion features for various vision tasks. Deep learning-based methods~\cite{li2021rfn},~\cite{tang2022image},~\cite{zhao2023metafusion} have advanced image fusion significantly. Through cross-view fusion of infrared and visible images, a fused result with abundant texture details and salient thermal targets can be generated.
Notably, external noise can impact fusion results. As shown in Fig.~\ref{fig1}, violated by rainfall degradation, the performance of existing advanced fusion algorithms is concerning. Therefore, we propose a novel cross-view sensor information cleaning framework for de-raining at nighttime. Concretely, we augment the de-raining process with cross-view sensor information fusion. This fusion provides a more comprehensive environmental representation and improves the visual perception under low-light conditions. Consequently, we produce de-rained results with rich detailed textures, enhanced contrast, and improved visual perception, contributing to the safety and reliability of nighttime driving.

In summary, the main contributions of this paper are as follows:

\begin{itemize}
    \item[$\bullet$] We first address image de-raining in low-light conditions for dynamic driving scenes. A novel cross-view sensor cooperative learning is proposed for image de-raining in low-light nighttime scenes.
    \item[$\bullet$] We design an information cleaning module, i.e., CleanNet, based on the convolutional nerual network (CNN)-Transformer structure to clean disruptive information such as rain streaks. This module effectively models and learns the global and local features of rainfall degradation, better aggregates the features and reconstructs rain-free images with high quality.
    \item[$\bullet$] A new cross-view sensor information fusion module,  i.e., FusionNet, is designed to fuse source information captured by sensors from different views. This module produces clean-fusion images with richer and clearer textures and more salient scenes. Simultaneously, through iterative refinement of the fused images, we achieve higher contrast and improved visual quality.
    \item[$\bullet$] Extensive experiments demonstrate that our proposed method offers an effective solution for challenging driving scenarios in rainy nighttime. It not only addresses the gap in existing de-raining methods for specific nighttime low-light scenarios, but also extends the applicability of de-raining and image fusion techniques in computer vision tasks.
\end{itemize}

\section{Related work}
\subsection{Image De-raining}
Single image de-raining has acquired substantial research interest in recent years, featuring traditional algorithms and deep learning methods.Early traditional techniques~\cite{zhang2017convolutional}, \cite{gu2017joint},~\cite{kang2011automatic},~\cite{li2016rain},~\cite{luo2015removing}, utilize the prior knowledge of rain streaks to realize de-raining.
Kang~\emph{et al.}~\cite{kang2011automatic} proposed the first automatic image decomposition framework guided by morphological component analysis (MCA), which transforms the problem of rainwater removal into an image decomposition problem leveraging MCA. Li~\emph{et al.} described rain streak removal as a layer decomposition task in~\cite{li2016rain}, where image patch-based priors were used to distinguish between the background layer and rain line layer.
Relying on Gaussian mixture models, Priors adapt to various directions and scales of rain marks. However, manually designed priors based on empirical statistical results struggle to model clean backgrounds and rain lines effectively in complex rainy images. Consequently, the effectiveness of traditional de-raining approaches remains significantly constrained.

Recently, with the superior performance of deep learning in image processing, approaches grounded in deep learning have been proposed for image restoration~\cite{jin2019flexible},~\cite{chen2022unpaired}~\cite{kulkarni2022unified}.
The deep CNN-based frameworks~\cite{jiang2020decomposition},~\cite{wang2020deep},~\cite{jiang2021rain},~\cite{yang2021end},~\cite{wang2023multi} have achieved considerable outcomes in image de-raining. Fu~\emph{et al.}~\cite{fu2017removing} introduced a deep detail network that utilized deep residual networks. Lin~\emph{et al.}~\cite{lin2020utilizing} integrated two-phase processing and a fuzzy broad learning system to form a novel framework for image de-raining. In~\cite{que2020attentive}, a unified framework based on an attentive composite residual network was proposed for single image-based rain removal. These CNN-based methods fully consider rain degradation characteristics, such as direction, scale, and density.
However, they are subject to the inherent constraints of local connectivity and translation invariance of convolution operations. Thus, there are limitations in mining global information and restoring delicate textural details.

The Transformer~\cite{dosovitskiy2020image} has achieved remarkable performance in natural language processing~\cite{vaswani2017attention}, and has also been explored in vision tasks~\cite{chen2020transformer},~\cite{zhou2023utlnet}. 
Methodologies~\cite{liang2022drt},~\cite{qin2021etdnet},~\cite{xiao2022image},~\cite{wang2022uformer},~\cite{zamir2022restormer} based on Transformer are progressively applied to various image reconstruction tasks, including image de-raining. 
Chen~\emph{et al.}~\cite{chen2021pre} introduced a network architecture grounded in pure Vision Transformer for handling image restoration tasks such as de-raining, denoising, and super-resolution. Xiao~\emph{et al.}~\cite{xiao2022image} presented a Transformer-based mirror de-raining architecture, efficiently achieving de-raining reconstructions through complementary window-based and spatial-based Transformers.
Leveraging the self-attention mechanism, Transformer is widely utilized to capture long-range dependencies in image restoration tasks. In image de-raining, both global dependencies and local features of raindrops are indispensable. Only by deeply extracting and aggregating these two features can higher-quality rain-free images be reconstructed. While Transformer-based techniques excel at modelling long-range dependencies, CNN-based methods are still preferred for extracting local features and information.

In this regard, hybrid architectures~\cite{jiang2022magic},~\cite{chen2023hybrid},~\cite{chen2023sparse} have emerged to boost image de-raining performance. Chen~\emph{et al.}~\cite{chen2023hybrid} also proposed a hybrid CNN-Transformer feature fusion network, leveraging the learning advantages of both architectures to achieve high-quality de-raining results.

In our method, inspired by DRSformer, we design a CNN-Transformer-based information cleaning module, CleanNet. This modular network aims to effectively remove interference factors, achieve improved de-raining performance, and generate clean outcomes with finer textures.

\subsection{Image Fusion}
Image fusion can effectively integrate merits from different source images, complementing each other to generate finer fused images. Therefore, multi-modal image fusion techniques are widely used across several computer vision domains~\cite{ha2017mfnet},~\cite{lu2020cross}. 

Traditional image fusion technologies~\cite{zhou2016perceptual},~\cite{li2020fast},~\cite{ma2020infrared} primarily focus on feature extraction and feature integration. Depending on the characteristics of the extracted features, different fusion rules are designed to accommodate various fusion tasks. Multi-scale transformations-based methods, such as Laplacian pyramid, discrete wavelet, discrete cosine, contourlet, and shearlet have been successfully integrated into image fusion frameworks~\cite{liu2014region}, \cite{bhateja2015multimodal},~\cite{yang2016multimodal},~\cite{chen2017fusion},~\cite{jiang2018multi} for feature extraction and reconstruction. 
Sparse presentation techniques~\cite{liu2016image},~\cite{wu2020infrared} represent image patches with over-complete dictionaries and corresponding sparse coefficients, fusing images by merging these coefficients.
subspace analysis methods like independent component analysis~\cite{cvejic2007region}, and principal component analysis~\cite{fu2016infrared} are also ultilized in image fusion, enhancing the intrinsic information within images by projecting high-dimensional images into lower-dimensional subspaces.
Optimization-based techniques~\cite{ma2016infrared} and hybrid methods~\cite{ganasala2019contrast} have also been applied in the field of image fusion. 
However, designing fusion rules that align with various feature extraction methods poses a key challenge in traditional image fusion, with manually crafted rules proving challenging to adapt to increasingly complex extraction techniques.

With the advancement of deep learning in computer vision tasks, deep learning-based methods have been applied in image fusion to tackle the challenges faced by traditional algorithms. 
PMGI~\cite{zhao2021efficient} presents a fast and unified image fusion framework, addressing various image fusion tasks end-to-end without crafting fusion rules manually.
In view of the cross-domain interaction between different fusion tasks, U2Fusion~\cite{xu2020u2fusion} trained a unified model with shared parameters that alleviated storage difficulty and catastrophic forgetting in continuous learning.
Tang~\emph{et al.}~\cite{tang2022image} proposed a real-time image fusion network combining image fusion with semantic segmentation. Recently, the Transformer architecture has also been applied to image fusion frameworks. Ma~\emph{et al.}~\cite{ma2022swinfusion} proposed a general fusion framework based on cross-domain remote learning and the Swin Transformer.

Furthermore, generative adversarial networks (GAN)-based methods employ adversarial loss functions to form unsupervised fusion frameworks. Ma~\emph{et al.} proposed FusionGAN~\cite{ma2019fusiongan}, utilizing a generative adversarial network to fuse visible and infrared images. Ma~\emph{et al.} further introduced the dual discriminator GAN called DDcGAN~\cite{ma2020ddcgan}, to balance thermal radiation and texture details in the fused image. While Li~\emph{et al.} introduced the AttentionFGAN~\cite{li2020attentionfgan}, integrating multi-scale attention mechanisms into both the generator and discriminator.

Nevertheless, these existing methods overlook the impact of external environment factors on multimodal sensor data. In scenarios with poor illumination, brightness information in visible images seriously degrades, leading to the deterioration or loss of texture structures. Tang~\emph{et al.} introduced PIAFusion~\cite{tang2022piafusion}, which addresses this issue by incorporating illumination perception into the fusion process.

In our method, we design a cross-view information fusion module, FusionNet, and introduce a cascaded refinement unit for balancing the image contrast and color. Within FusionNet, we seamlessly integrate multi-level features from de-rained visible and infrared images. Through iterative refinement of the clean-fused image, we attain final de-rained results with markedly enhanced contrast and superior visual quality.

\section{Observation}
Generally, image de-raining and image fusion methods often assume that the operating environment is ideal and use pure data for implementation, which is not bound by real-world situations. However, interference in real-world environments often diminishes the contrast of scenes within the driving line of sight. Based on this motivation, existing image de-raining methods are first re-examined with extreme yet practical conditions as follows:

\begin{itemize}
    \item[$\bullet$] \textbf{Low-light condition.} As shown in Fig.~\ref{fig2}, existing image de-raining methods exhibit limitations in low-light nighttime scenarios. In daytime scenes, typical de-raining algorithms are generally effective in removing rainwater noise. However, in nighttime scenes, visible-light cameras fail to support the de-raining algorithm with poor illumination. As illustrated in Fig.~\ref{fig2}, in nighttime driving scenarios with heavy rain, we observed that current de-raining approaches often fail to restore structure-preserving scenes and highlight targets. To this end, existing de-raining algorithms face challenges in assisting automatic systems and capturing surroundings in low-light conditions.
    
    \item[$\bullet$] \textbf{Cross-sensor information fusion.} In autonomous vehicles, multi-sensor (for example, RGB and infrared sensors) is a standard configuration. Images captured from different sensors, such as visible and infrared (IR) images, have complementary information and hold the potential for significant improvements in nighttime de-raining tasks. While IR images remain unaffected by lighting and provide clearer target scenes, they lack color and texture details crucial for human perception. Visible RGB images, As shown in Fig.~\ref{fig3}, offer higher spatial resolution, detailed textures, and better contrast compared to IR images. However, IR images excels in reflecting thermal targets even with poor illumination. Cross-sensor information fusion renders fused images capable of preserving detailed textures and thermal information from various inputs. Nighttime de-raining tasks can benefit from these fusion images, thus enhancing the safety of real-world driving scenes.
    
    \item[$\bullet$] Based on the above observations, we aim to address the challenge of nighttime image de-raining by employing cross-view sensor information fusion. We design a two-stage framework for information cleaning and fusion. The first-stage network performs information cleaning on the rainy-visible images and reconstructs clean outputs. These clean-visible images are subsequently fused with their corresponding infrared images across sensors in the second stage. As depicted in Fig.~\ref{fig4}, it is obvious that our method not only effectively removes rain streaks, but also maintains the merits of both visible and infrared images. Our results restore detailed textures and prominent targets, with significant overall brightness and contrast.
\end{itemize}

\begin{figure}[!t]
\centering
\includegraphics[width=0.99\columnwidth]{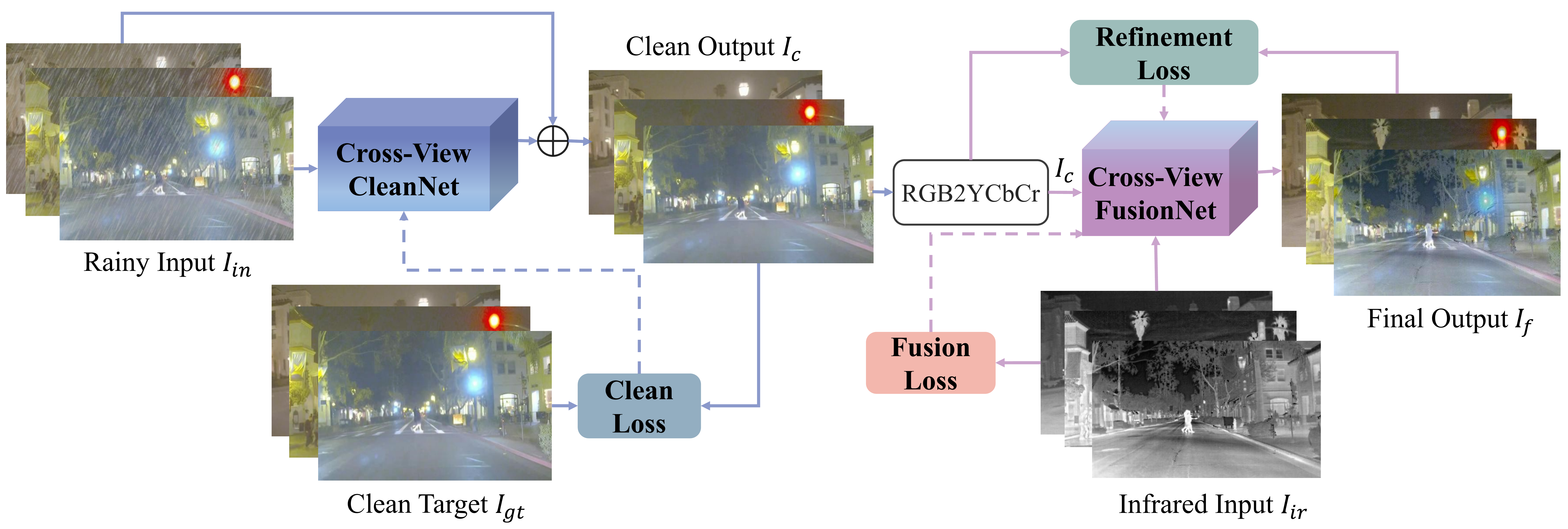}
\caption{The overall framework of the proposed CVCL for nighttime image de-raining. Cross-view CleanNet comprises the first stage network, while cross-view FusionNet comprises the second stage network. CleanNet reconstructs high-quality de-rained results. FusionNet fully integrates the clean-visible and infrared images to generate final clean-fusion results. The symbol $\oplus$ represents the sum operation.}
\label{fig5}
\vspace{-3mm}
\end{figure}

\begin{figure*}[!t]
\centering
\includegraphics[width=0.95\linewidth]{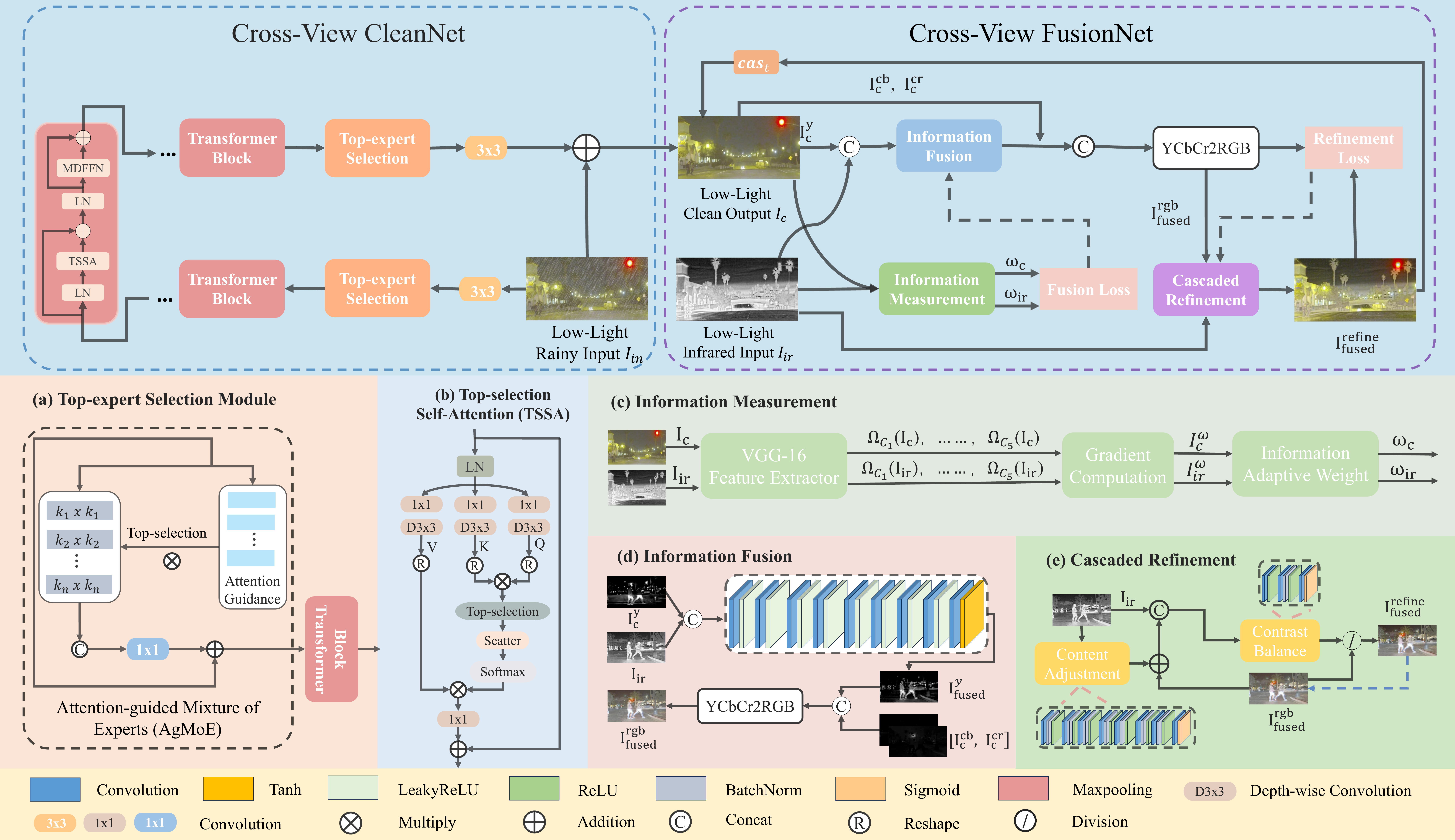}
\caption{The overall architecture of the proposed two-stage CVCL networks. The pipeline in Cross-View CleanNet aims to remove rain streaks and reconstruct clean results, which mainly consists of attention-guided mixture of experts (AgMoE), Transformer block with top-selection self attention (TSSA) and mutual deformable feed-forward network (MDFFN). $I_{in}$ denotes the low-light rainy input, and LN signifies the layer normalization. The Cross-View FusionNet is designed to integrate comprehensive information for nighttime driving, consisting of information measurement and fusion. $I_c$ denotes the low-light clean output from the CleanNet stage, and $I_{ir}$ corresponds to the infrared image captured by the infrared sensors in rainy scenarios.}
\label{fig6}
\vspace{-3mm}
\end{figure*}

\section{Methodology}
\label{sec:guidelines}
In this section, we first introduce the entire workflow of our proposed nighttime image de-raining with cross-view sensor information fusion. Then, we describe the details of each module.

\subsection{Overall Framework}
The overall framework of our proposed CVCL method for nighttime image de-raining is shown in Fig.~\ref{fig5}. In the overall working architecture, we employ a two-stage structure.

\begin{itemize}
    \item[$\bullet$] In the first stage, we design an information cleaning module (i.e., CleanNet). This module facilitates a feature extraction toward intricate rainfall degradation for reconstructing higher-quality de-rained images. Given a low-light rainy image $I_{in} \in R^{H \times W \times 3}$, CleanNet extracts both local features and long-range dependencies in rainfall degradation, yielding clear rain-free results, denoted as $I_c \in R^{H \times W \times 3}$.
    
    \item[$\bullet$] In the second stage, we design a cross-view sensor information fusion module (i.e., FusionNet). This network fully integrates complementary information from visible and infrared images to produce final clean-fusion images with richer information, enhanced contrast and better visual quality. Specifically, FusionNet conducts cross-sensor information fusion on $I_c$ and infrared images $I_{ir} \in R^{H \times W \times 1}$ to achieve high-quality fusion results.
\end{itemize}

\subsection{Cross-View CleanNet}
For the visible inputs, CleanNet restores scenes obscured by rainwater and reconstructs high-quality de-rained results. The overall structure of CleanNet is shown in Fig.~\ref{fig6}. CleanNet includes the attention-guided mixture of experts (AgMoE), top-selection self attention (TSSA), and mutual deformable feed-forward network (MDFFN).
    
\subsubsection{Attention-guided mixture of experts (AgMoE)}
To better extract the local features of spatially varying rainwater distribution, we construct an AgMoE. Specifically, we employ multiple convolution operations to form parallel expert layers, including average pooling with a receptive field of 3$\times$3, separable convolution layers with kernel sizes of 1$\times$1, 3$\times$3, 5$\times$5, and 7$\times$7, and dilated convolutional layers with kernel sizes of 3$\times$3, 5$\times$5, and 7$\times$7. Unlike the traditional standard mixture of experts, we do not use gating networks to work in conjunction with experts. Instead, we introduce a self-attention mechanism to assign weights to each expert module.

Given an input feature map $F_{n-1} \in R^{H \times W \times  C}$, channel-wise averaging is first applied to generate a description $x_c \in R^c$ as:
\begin{equation}
    x_c = \frac{1} {H \times W} \sum_{i=1}^{H} \sum_{j=1}^{W} F_{n-1}(i,j)
\end{equation}
where $F_{n-1}(i,j)$ represents the $(y,x)$ position of the feature map $F_{n-1}$. Each expert receives a self-attention weight vector from its corresponding learnable attention matrices $W_1\in R^{M \times C}$ and $W_2\in R^{N \times M}$. Here, $M$ is the dimension of the weight matrices. The feature representation extracted through the AgMoE module, denoted as $Z_f$, is expressed as follows:
\begin{equation}
    Z_f = [h_{expert}^S(W_2 \sigma (W_1 x_c))]
\end{equation}
where $h_{expert}$ represents the expert layer and S denotes the number of experts. $\sigma(\cdot)$ is the ReLU function, and $[\cdot]$ represents channel-wise concatenation. Between each AgMoE module, skip connections are also used to connect continuous intermediate features to ensure stable training. The final output of the \emph{n}-th AgMoE module is given by:
\begin{equation}
    F_n = f_{1x1}(Z_f)+F_{n-1}.
\end{equation}
Here, $f_{1x1}(\cdot)$ represents a convolutional layer with \emph{C} filters.

\subsubsection{Top-selection self attention (TSSA)}
We introduce the top-selection algorithm to construct a novel TSSA to capture long-range dependencies within global features more effectively. As shown in Fig.~\ref{fig6}, the TSSA operations are introduced as follows:
\begin{equation}
    \begin{split}
        f_{TS}(F_n) &= f_{scatter}(f_{top-k}(F_n)),\\
        f_{TSSA}(F_n) & = softmax(f_{TS}(\frac{QK^T} {\sqrt{d}}))V.
    \end{split}
\end{equation}
Here, $F_n$ refers to the features obtained from the AgMoE, $Q$, $K$, and $V$ respectively represent the query, key and value with dimensions $R^{I \times d}$. $f_{top-k}(\cdot)$ indicates retaining the top-k elements with higher attention weights using the top-selection algorithm. $f_{scatter}(\cdot)$ uses the scatter function to replace the normalized scores of elements below the top-k scores with zero.
    
\subsubsection{Mutual deformable feed-forward network (MDFFN)}
\emph{In contrast to conventional feed-forward networks, we construct a new MDFFN.} It employs two parallel depth-wise convolution branches, with convolution kernels of 3$\times$3 and 5$\times$5 respectively. MDFFN achieves multi-scale feature representations by cross-scale feature fusion via channel-wise concatenation, enhancing the extraction of rainfall degradation features and promoting rainwater removal.

Given the features $F_n$ obtained from the AgMoE, the cascaded encoded features can be acquired through the following process:
\begin{equation}
    \begin{split}
        F_n^{'} &= F_n + f_{TSSA}(f_{LN}(F_n)),\\
        F_{n+1} &= F_n^{'} + f_{MDFFN}(f_{LN}(F_n^{'})).
    \end{split}
\end{equation}
Here, $f_{TSSA}$ and $f_{MDFFN}$ represent TSSA and MDFFN, respectively. $f_{LN}$ denotes layer normalization.

\subsection{Cross-View FusionNet}\label{formats}
The clean-visible images $I_c$ from CleanNet and their corresponding infrared channel images $I_{ir}$ serve as the RGB-IR source inputs for FusionNet. As shown in Fig.~\ref{fig6}, FusionNet contains two modules: information measurement and fusion. 

\subsubsection{Information Measurement and Fusion}
To preserve the critical information, we use an information measurement strategy. The clean-visible image is converted from RGB into YCbCr, which yields three separate channel images, $I_c^y$, $I_c^{cr}$, and $I_c^{cb}$. As the luminance channel contains more promonent structural details and brightness variations than the chrominance channel, we choose the luminance channel for fusion.

We use the pre-trained model of VGG-16~\cite{simonyan2014very} to extract multi-level features of $[I_c^y,I_{ir}]$. Shallow features $\Omega_{C_1}(I)$, $\Omega_{C_2}(I)$, and $\Omega_{C_3}(I)$ capture texture and shape details, while high-level feature maps $\Omega_{C_4}(I)$ and $\Omega_{C_5}(I)$ reveal deep features such as the image content and spatial structure. Information measurement is performed on these extracted feature maps to obtain corresponding values $I_c^w$ and $I_{ir}^w$ as: 
\begin{equation}
    I^w = \frac{1} {N} \sum_{k=1}^{G} \frac{1} {H_k W_k D_k}\sum_{n=1}^{D_k} \| \nabla\Omega_{C_n^k}(I) \|_F^2.
\end{equation}
Here, $k$ represents the \emph{k}-th max-pooling layer. $H_k$, $W_k$, and $D_k$ denote the height, width, and number of channels of the feature in the k-th layer respectively. $\nabla$ signifies the Laplacian operator, and $\| \cdot \|_F$ indicates the Frobenius norm. We introduce two adaptive weights, $\omega_c$ and $\omega_{ir}$, as the degree of information preservation. These weights are determined by the information measurement values, defined as follows:
\begin{equation}
    [\omega_c, \omega_{ir}] = softmax(I_c^w, I_{ir}^w)
\end{equation}
where $I_c$ and $I_{ir}$ represent the clean-visible image and infrared image respectively. Through the Softmax function, the weights are normalized to [0, 1], with their sum equal to 1. Then, $\omega_c$ and $\omega_{ir}$ are utilized in the fusion loss function to control the retention of different source information.

The paired $I_c^y$ and $I_{ir}$ are concatenated and fed into the information fusion module to generate the fused image $I_{fused}^y$, defined as follows:
\begin{equation}
    I_{fused}^y = F(\theta, I_c^y, I_{ir}).
\end{equation}
Here, \emph{F} represents the information fusion module, and $\theta$ represents the corresponding parameters.

\subsection{Cascaded Refinement for Nighttime De-raining}
To further improve the contrast and color balance of the fused image, we incorporate a cascaded refinement module for self-supervised optimization. It includes content adjustment and contrast balance structures. The fused image iteratively interacts with the infrared content to obtain contrast information, refining the image for contrast balance and enhancement. This iterative process is conducted in multiple cascaded stages. Each level of refined-fused image serves as the visible input for the next level of fusion, thus achieving a cascaded refinement. The formula is as follows:
\begin{equation}
    I_{fused}^{t+1} = \frac{I_{fused}^{t}}{B_t[(I_{fused}^{t} + A_t(I_{ir}))\otimes I_{ir}]}.
\end{equation}
Here, $A(\cdot)$ signifies Content Adjustment, $B(\cdot)$ represents Contrast Balance, and $\otimes$ denotes concatenation. $A(\cdot)$ comprises five 3$\times$3 convolutional layers, with the first four followed by batch noemalization and ReLU activation function, and the last layr followed by a Sigmoid activation function. $B(\cdot)$ include three 3$\times$3 convolutional layers, with the first two using ReLU activation fuction and the last employing a Sigmoid function as activation. $T=3$ indicates the number of adjustment iterations per stage, and $I_{fused}^{t}$ is the output of multiple refinement iterations within one stage.

\begin{table*}[t]
    \caption{The loss functions employed in the proposed framework.}
    \label{table1}
    \centering
    \small
    \resizebox{2\columnwidth}{!}{
    \begin{tabular}{|c|c|c|c|}
    \hline 
        \thead{Modular Network} & Loss Function & Composition & Object \\ \hline  \hline
        CleanNet & Clean Loss  & $L_1$ regularization term & \thead{Clean output $I_c$ and ground-truth $I_{gt}$: \\ $L_1$($I_c$, $I_{gt}$)} \\ \hline
        \multirow{6}{*}{\thead{FusionNet}} 
        & \multirow{3}{*}{\thead{Fusion Loss}} & \thead{Structural similarity loss} & \multirow{2}{*}{\thead{Clean-visible image $I_c$, infrared image $I_{ir}$, and fused image $I_{fused}$: \\ $\mathcal{L}_{fuse}$($I_c^y$, $I_{ir}$, $I_{fused}^y$)}} \\
        \cline{3-3}
        \multirow{4}{*}{} & \multirow{2}{*}{} & \thead{Mean square error loss} & \multirow{2}{*}{} \\
        \cline{2-4}
        \multirow{4}{*}{} & \multirow{3}{*}{\thead{Refinement Loss}} & \thead{Consistency loss} & \multirow{2}{*}{\thead{Fused image $I_{fused}$ and fused-refine image $I_{fused}^{refine}$: \\ $\mathcal{L}_{refine}$($I_{fused}$, $I_{fused}^{refine}$)}} \\
        \cline{3-3}
        \multirow{4}{*}{} & \multirow{2}{*}{} & \thead{Smoothness loss} & \multirow{2}{*}{} \\ 
    \hline
    \end{tabular}
    }
    \label{tab1}
\end{table*}

\begin{algorithm}[!ht]
    \renewcommand{\algorithmicrequire}{\textbf{Input:}}
	\renewcommand{\algorithmicensure}{\textbf{Output:}}
	\caption{Cross-View Sensor Cooperative Learning Algorithm}
	\label{power}
    \begin{algorithmic}
        \REQUIRE  rainy-visible data $I_{in}$, target visible data $I_{gt}$, and infrared data $I_{ir}$
	    \ENSURE clean fusion result $I_f$ (final fusion output)
	    
	    \STATE $\triangleright$ Information Cleaning
	    \FOR {$n$ $\leq$ cleaning training iters $N$}
	        \STATE Update the clean loss according to $\mathcal{L}_{clean} = \| I_c - I_{gt} \|_1$
	        \STATE Update the parameters of CleanNet $\Phi_C$ by AdamW Optimizer: $\nabla_{\Phi_C}$($\mathcal{L}_{clean}(\Phi_C)$)
	    \ENDFOR
	    \STATE $\triangleright$ Cross-View Sensor Fusion
	    \FOR {$e$ $\leq$ fusion training epochs $E$}
	        \STATE Generate clean-visible images $I_c$ with cleaning model in the training set and construct                					clean-visible and infrared pairs $\left\{I_c, I_{ir} \right\}$
	        \FOR{$k$ $\leq$ cascaded stages $K$}
	            \STATE Convert the clean-visible images $I_c$ to the YCbCr space and extract the Y channel $I_c^y$ from $I_c$
	            \STATE Update the fusion loss according to $\mathcal{L}_{fuse}$ = $\mathcal{L}_{ssim}$ + $\alpha$ * $\mathcal{L}_{mse}$
	            \STATE Update the parameters of the fusion module $\Phi_F$ by Adam Optimizer: $\nabla_{\Phi_F}$($\mathcal{L}_{fuse}(\Phi_F)$)
	            \STATE Obtain the fused images: $I_{\text{fused}}^{\text{rgb}} \leftarrow \{I_{\text{fused}}^y, I_c^{\text{cr}}, I_c^{\text{cb}}\}$
	            \FOR{$t$ $\leq$ adjustment iterations $T$}
	            \STATE Update the refinement loss according to $\mathcal{L}_{refine}$ = $\mathcal{L}_{smo}$ + $\beta$ * $\mathcal{L}_{con}$
	            \STATE Update the parameters of the cascaded refinement module $\Phi_R$ by Adam Optimizer: $\nabla_{\Phi_R}$($\mathcal{L}_{refine}(\Phi_R)$)
	            \ENDFOR
	            \STATE $I_{fused}^{rgb}$ = $I_{fused}^{refine}$
	        \ENDFOR
	    \ENDFOR
    \end{algorithmic}
\end{algorithm}

\subsection{Loss Function}
We design specific loss functions for each modular network, as shown in Table~\ref{tab1}. Guided by these loss functions, our model is capable of generating high-quality clean-enhanced images with intricate details and significant contrast.

\subsubsection{Clean Loss}
To ensure the effectiveness of CleanNet in removing rain streaks and reconstructing high-quality clean images, we optimize the model by minimizing the following clean loss function:
\begin{equation}
    \mathcal{L}_{clean} = \| I_c - I_{gt} \|_1.
\end{equation}
Here, $I_c=\phi(I_{in})+I_{in}$, represents the clean output of CleanNet, and $\phi(\cdot)$ denotes CleanNet. $I_{gt}$ represents the ground-truth clean image, and $\| \cdot \|_1$ signifies the $L_1$-norm.

\begin{figure*}[!t]
\centering
\includegraphics[width=0.95\linewidth]{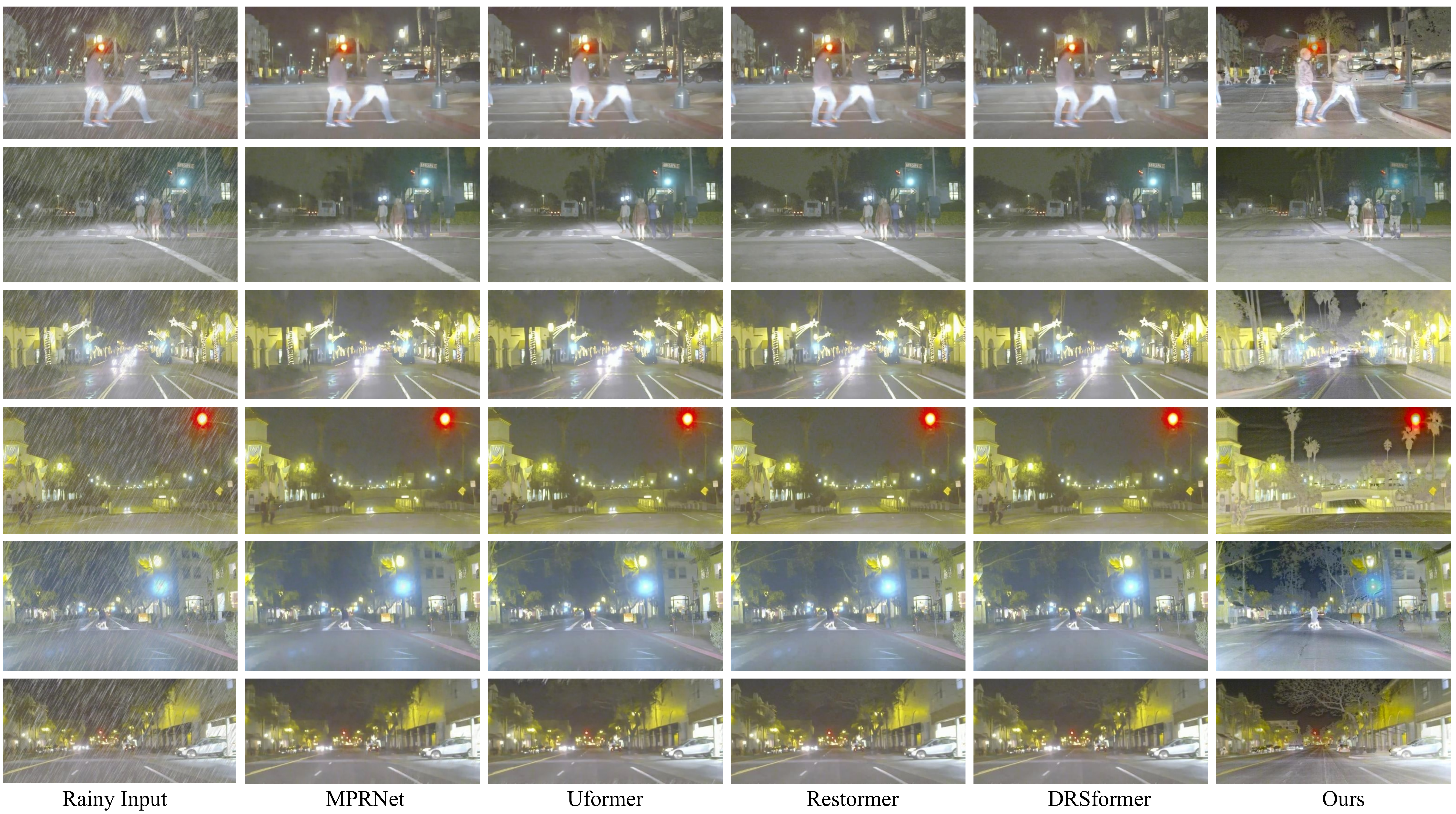}
\caption{Visual qualitative comparison of our method with four SOTA de-raining methods on six scenes from RoadScene-rain dataset. For clearer comparison, please zoom in the figures.}
\label{fig7}
\vspace{-3mm}
\end{figure*}

\begin{figure}[!t]
\centering
\includegraphics[width=0.95\columnwidth]{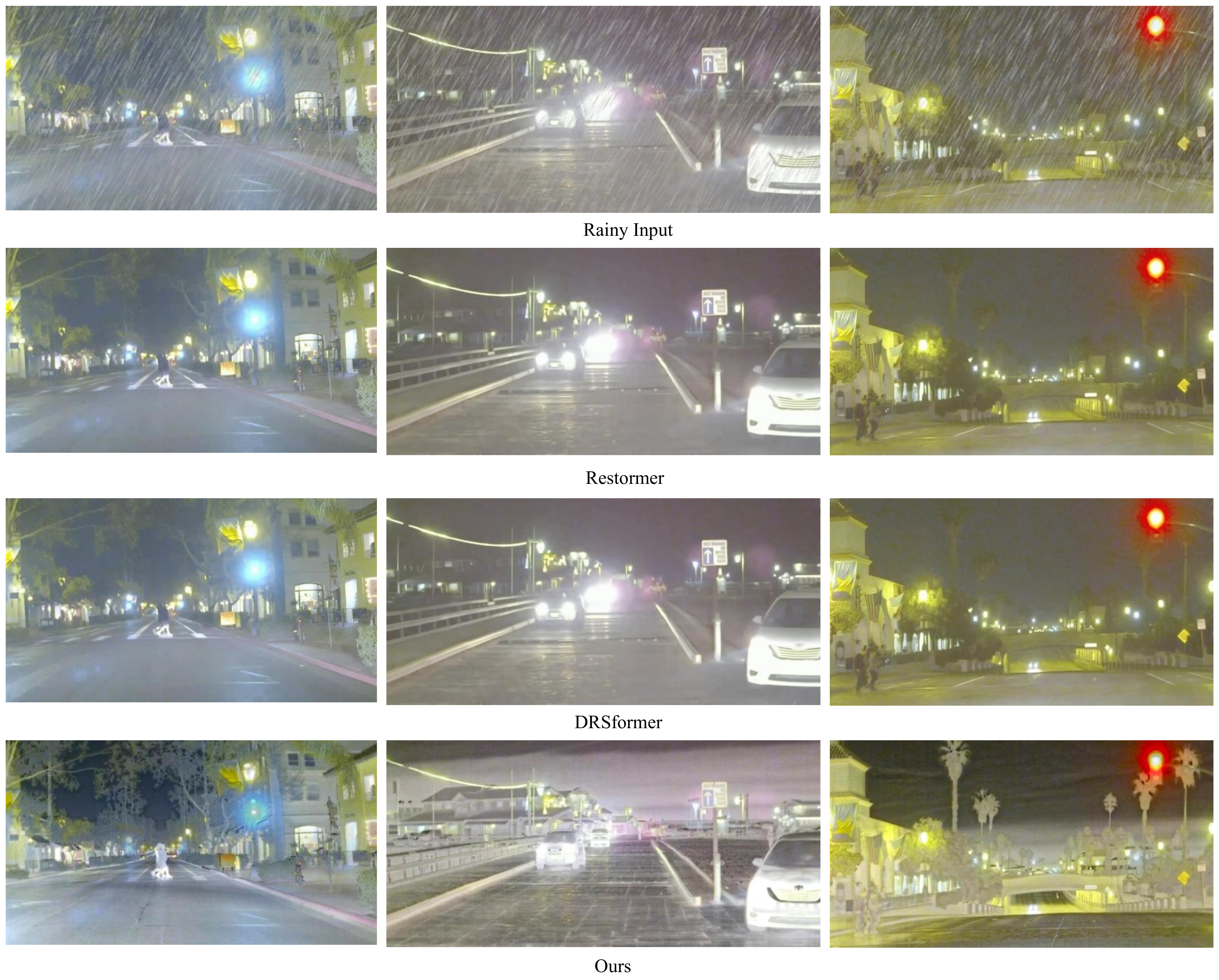}
\caption{Intuitive qualitative comparison results of our method with some de-raining algorithms on three scenes from RoadScene-rain dataset.}
\label{fig8}
\vspace{-3mm}
\end{figure}

\begin{figure}[!t]
\centering
\includegraphics[width=0.95\columnwidth]{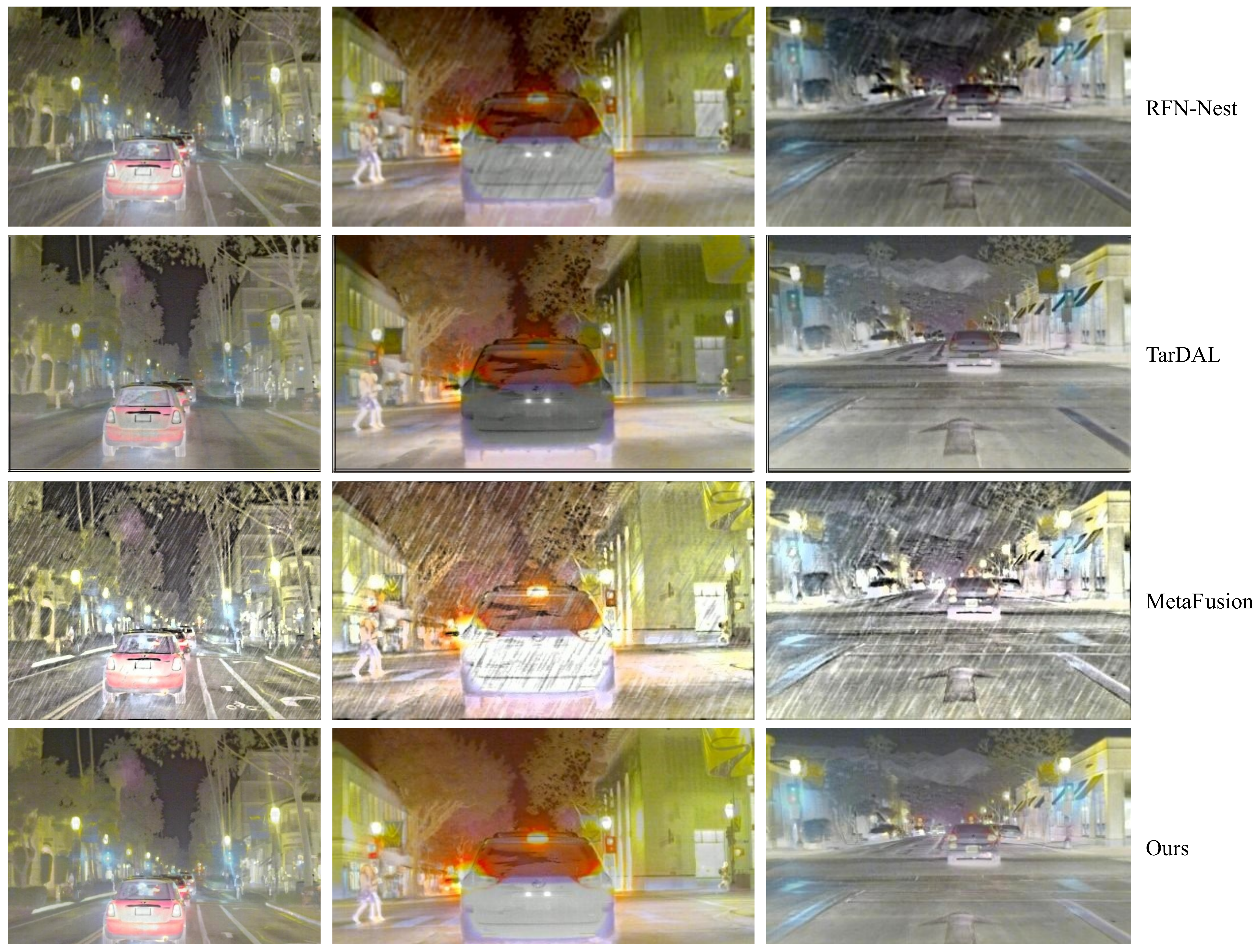}
\caption{Intuitive qualitative comparison results of our method with some fusion algorithms on three scenes from RoadScene-rain dataset.}
\label{fig9}
\vspace{-3mm}
\end{figure}

\subsubsection{Fusion Loss}
We expect that FusionNet can produce clean-fused images with abundant details and heightened structure fidelity. Thus, the fusion loss is designed as follows:
\begin{equation}
    \mathcal{L}_{fuse} = \mathcal{L}_{ssim} + \alpha * \mathcal{L}_{mse}
\end{equation}
where $\mathcal{L}_{ssim}$ and $\mathcal{L}_{mse}$ represent the structural similarity loss and the mean square error loss, respectively. $\alpha$ serves as a hyperparameter used to achieve a balance between the two loss components.

The structural similarity (SSIM) is a metric that evaluates distortion based on the luminance, contrast, and structural information similarity. We ultilize the structural similarity loss to constrain structural similarity, defined as follows:
\begin{equation}
    \begin{split}
       \mathcal{L}_{ssim} = \omega_c * (1 - SSIM(I_c^y, I_{fused}^y))\\
       +\ \omega_{ir} * (1 - SSIM(I_{ir}, I_{fused}^y)).
    \end{split}
\end{equation}
The mean squared (MSE) loss is used to constrain the difference in intensity distribution, defined as follows:
\begin{equation} 
    \begin{split}
        \mathcal{L}_{mse} = \omega_c * MSE(I_c^y, I_{fused}^y)\\
        +\ \omega_{ir} * MSE(I_{ir}, I_{fused}^y).
    \end{split}
\end{equation}
Here, $I_c^y$, $I_{fused}^y$, and $I_{ir}$ respectively correspond to the Y channel of the de-rained visible image, Y channel of the fused image, and infrared image. $\omega_c$ and $\omega_{ir}$ respectively represent the weight coefficients of the de-rained visible image and the infrared image.

To improve the visual quality and contrast of images generated by FusionNet, we design a novel refinement loss. It comprises two components, consistency loss and smoothness loss. The refinement loss is defined as follows:
\begin{equation}
    \mathcal{L}_{refine} = \mathcal{L}_{smo} + \beta * \mathcal{L}_{con}
\end{equation}
where $\mathcal{L}_{con}$ denotes the consistency loss, and $\mathcal{L}_{smo}$ signifies the smoothness loss, as proposed by~\cite{guo2020zero}. $\beta$ is a hyperparameter that balances these two components.

\begin{table*}[t]
    \caption{Quantitative results of our method with seven SOTA fusion methods on 40 image pairs from RoadScene-rain dataset. The best result is denoted by red, and the second result is indicated by blue.}
    \label{table2}
    \small
    \centering
    \resizebox{2\columnwidth}{!}{
    \begin{tabular}{|c|c|c|c|c|c|c|c|c|c|c|c|}
    \hline
        \thead{Metrics} & PSNR & SSIM & MS-SSIM & MI & VIF & CC & FMI\_pixel & FMI\_dct & FMI\_w & MSE & Qabf \\ \hline \hline
        \thead{PFGAN} & 61.8109 & 0.0756 & 0.5074 & 2.3273 & 0.1567 & 0.4149 & 0.8059 & 0.1226 & 0.1965 & 0.0498 & 0.1980  \\ \hline
        \thead{U2Fusion} & \textcolor{blue}{63.6591} & 0.3915 & 0.8329 & 2.3417 & 0.4116 & 0.6044 & 0.8456 & 0.2671 & 0.3109 & \textcolor{blue}{0.0343} & 0.4100 \\
        \hline
        \thead{RFN-nest} & 61.8966 & 0.3623 & \textcolor{blue}{0.8735} & 2.6254 & 0.4451 & \textcolor{blue}{0.6271} & \textcolor{blue}{0.8557} & 0.1649 & 0.2361 & 0.0468 & 0.3645 
        \\
        \hline
        \thead{TarDAL} & 62.0222 & \textcolor{blue}{0.3967} & 0.7204 & \textcolor{red}{4.1283} & \textcolor{red}{0.6368} & 0.4406 & 0.8433 & 0.2161 & \textcolor{blue}{0.3423} & 0.0510 & \textcolor{blue}{0.4591} 
        \\
        \hline
        \thead{AUIF} & 61.5385 & 0.3607 & 0.8228 & 2.5161 & 0.4232 & 0.6054 & 0.8356 & \textcolor{blue}{0.2835} & 0.3160 & 0.0487 & 0.3886
        \\
        \hline 
        \thead{PIAFusion} & 62.2806 & 0.3103 & 0.7003 & 2.4122 & 0.3771 & 0.5622 & 0.8506 &  0.2015 & 0.2564 & 0.0476 & 0.4034
        \\
        \hline
        \thead{MetaFusion} & 61.4239 & 0.2803 & 0.7145 & 1.7981 & 0.3552 & 0.5700 & 0.8231 & 0.2329 & 0.2669 & 0.0530 & 0.3139
        \\
        \hline \hline
        \thead{Ours} & \textcolor{red}{64.7128} & \textcolor{red}{0.4909} &
        \textcolor{red}{0.8885} & \textcolor{blue}{2.9556} & \textcolor{blue}{0.5766} & \textcolor{red}{0.6501} & \textcolor{red}{0.8620} & \textcolor{red}{0.3632} & \textcolor{red}{0.3946} & \textcolor{red}{0.0275} & \textcolor{red}{0.4597} 
        \\
    \hline
    
    \end{tabular}
    }
    \label{tab2}
\end{table*}

\section{Experimental Validation}
In this section, we present our experimental dataset, training details, and comparative evaluations. We demonstrate the superiority of our proposed method through quantitative and qualitative assessments, along with ablation experiments to highlight the effectiveness of each module.

\subsection{Experiment setup}
\subsubsection{RoadScene-rain Dataset}
In our experiments, we use the RoadScene dataset of FLIR videos~\cite{xu2020u2fusion} for evaluation. This dataset accurately aligns RGB-IR images and contains diverse scenes including roads, vehicles, and pedestrians, effectively simulating low-light nighttime driving scenarios. To simulate rainy conditions, we employ a raindrop synthesis algorithm~\cite{choi2022synthesized} on RGB images within the RoadScene dataset. The resulting dataset is named the RoadScene-rain dataset, comprising 221 image pairs and accurately simulating the low-light rainy nighttime driving scenarios in the real world. We reserve 40 pairs as test samples and apply the remaining pairs for training.

\subsubsection{Evaluation metrics}
We use eleven metrics to quantitatively evaluate the fusion results: mutual information (MI), visual information fidelity (VIF), peak signal-to-noise ratio (PSNR), structural similarity (SSIM), mean square error (MSE), correlation coefficient (CC), pixel feature mutual information (FMI\_pixel), discrete cosine feature mutual information (FMI\_dct), wavelet feature mutual information (FMI\_w), and gradient-based fusion performance (Qabf).
These metrics serve different purposes in evaluating the fused images. MI measures information content from an information theory perspective. VIF evaluates visual fidelity from the perspective of human visual perception. PSNR assesses effective information, while SSIM compares the structural similarity. MSE quantifies differences, and CC reflects the linear correlation between source input and fused result. Qabf, FMI\_pixel, FMI\_dct, and FMI\_w estimate source information representation, offering a comprehensive image quality assessment. Except for MSE, a higher value of each metric indicates better quality in the fused image.

\begin{table*}[t]
    \caption{Ablation experiment results on 40 image pairs from RoadScene-rain dataset. The best result is denoted by red.}
    \label{table3}
    \centering
    \scriptsize
    \resizebox{2\columnwidth}{!}
    {
    \begin{tabular}{|c|c|c|c|c|c|c|c|}
    \hline
        \thead{Method} & SSIM & MI & VIF & Qabf & FMI\_dct & FMI\_w & SCD \\ \hline \hline
        \thead{ w/ TSSA + MDFFN} & 0.4898 & 2.9553 & 0.5763 & 0.4596 & 0.3628 & 0.3941 & 1.5728 \\ \hline
        \thead{ w/ AgMoE + MDFFN} & 0.4793 & 2.7992 & 0.5434 & 0.4421 & 0.3273 & 0.3663 & 1.5714 \\ \hline
        \thead{ w/ AgMoE + TSSA} & 0.4769 & 2.7979 & 0.5417 & 0.4397 & 0.3253 & 0.3641 & 1.5670
        \\ 
        \hline \hline
        \thead{ w/ Cascaded Refinement} & 0.4796 & 2.7994 & 0.5436 & 0.4420 & 0.3276 & 0.3666 & 1.5722 \\
        \hline
        \thead{ w/ Information Measurement} & 0.4641 & 2.9102 & 0.5435 & 0.3974 & 0.2775 & 0.3353 & 1.3782
        \\
        \hline \hline
        \thead{Full Model} & \color{red}0.4909 & \color{red}2.9556 & \color{red}0.5766 & \color{red}0.4597 & \color{red}0.3632 & \color{red}0.3946 & \color{red}1.5740
        \\
    \hline
    \end{tabular}
    }
    \label{tab3}
\end{table*}
\subsubsection{Implementation details}
The training details of our proposed framework is showm in Algorithm \ref{power}. For AgMoE, the initial channel \emph{C} is set to 48, and \emph{M} = 32 for the learnable weight matrix, while the number of experts \emph{S} is 8. The training samples are randomly cropped into 64 $\times$ 64 patches. The batch size is set to 8 for 300K iterations using the AdamW optimizer. The initial learning rate is set to 1e-4. We train the fusion network for 500 epochs using the Adam for optimization. The initial learning rate is set to 1e-4 for the information fusion and 3e-4 for the cascaded refinement. The weight decay of the optimizer is set to 3e-4. The number of cascaded stages \emph{K} = 3. For the hyperparameters in the loss functions, $\alpha$ and $\beta$ are set to 20 and 1.5. 

\subsection{Comparison with state-of-the-art methods}
We compare our proposed method with state-of-the-art (SOTA) methods in both image de-raining and image fusion. The comparison includes advanced image de-raining algorithms such as MPRNet \cite{zamir2021multi}, Uformer \cite{wang2022uformer}, Restormer \cite{zamir2022restormer}, and DRSformer \cite{chen2023learning}, as well as image fusion methods, such as PFGAN \cite{fu2021image}, AUIF \cite{zhao2021efficient}, RFN-nest \cite{li2021rfn}, U2Fusion \cite{xu2020u2fusion}, PIAFusion \cite{tang2022piafusion}, TARDAL \cite{liu2022target}, and MetaFusion \cite{zhao2023metafusion}. 

\subsubsection{Qualitative Results}
Fig.~\ref{fig7} and~\ref{fig8} show the visual results of our method in comparison with some SOTA image de-raining algorithms. It can be observed that image texture and scene details suffer in rainy and nighttime conditions. While existing de-raining methods like MPRNet, Uformer, Restormer and DRSformer effectively remove rain streaks, they still produce blurry textures and rough details, and targets in dark regions are difficult to discern. In comparison, our method excels in cleaning raindrops and restoring detailed textures in visible images. Our results offer richer textures, highlighted targets, brighter scenes, and enhanced contrast. Moreover, our results offer valuable road and environmental information, providing superior visual effects overall.

Fig.~\ref{fig9} illustrates visual comparisons between our approach and some SOTA image fusion methods. Obviously, our approach effectively removes raindrops, resulting in more detailed textures, sharper edges, and highlighted foreground targets even in dark regions. In comparison, other fusion methods still produce rainwater artifacts. Moreover, they struggle with edge artifacts, blurry details, and information loss due to poor illumination.

\subsubsection{Quantitative Results}
The comparison results of different methods across pixel-based metrics are presented in Table~\ref{tab2}. 
Our method outperforms others in several metrics on the RoadScene-rain dataset, including PSNR, SSIM, MS-SSIM, CC, FMI\_pixel, FMI\_dct, FMI\_w, Qabf, and MSE. The highest PSNR indicates that our method generates better-quality images with less noise. The best SSIM and MS-SSIM suggest that our results retain more structural information. The best result in CC implies a better preservation of structure and brightness information in our results. The outstanding performance in terms of the FMI\_pixel, FMI\_dct, FMI\_w, and MSE further demonstrates that our method provides higher fusion quality and fidelity. Additionally, our method ranks first in the Qabf metric, signifying effective edge information preservation. Furthermore, our method is only marginally behind TarDAL on the VIF metric, which means that our fusion results also have a good visual perception and are consistent with the human visual system. To this end, our method has obvious advantages compared to other SOTA methods. 

\begin{figure}[!t]
\centering
\includegraphics[width=0.95\columnwidth]{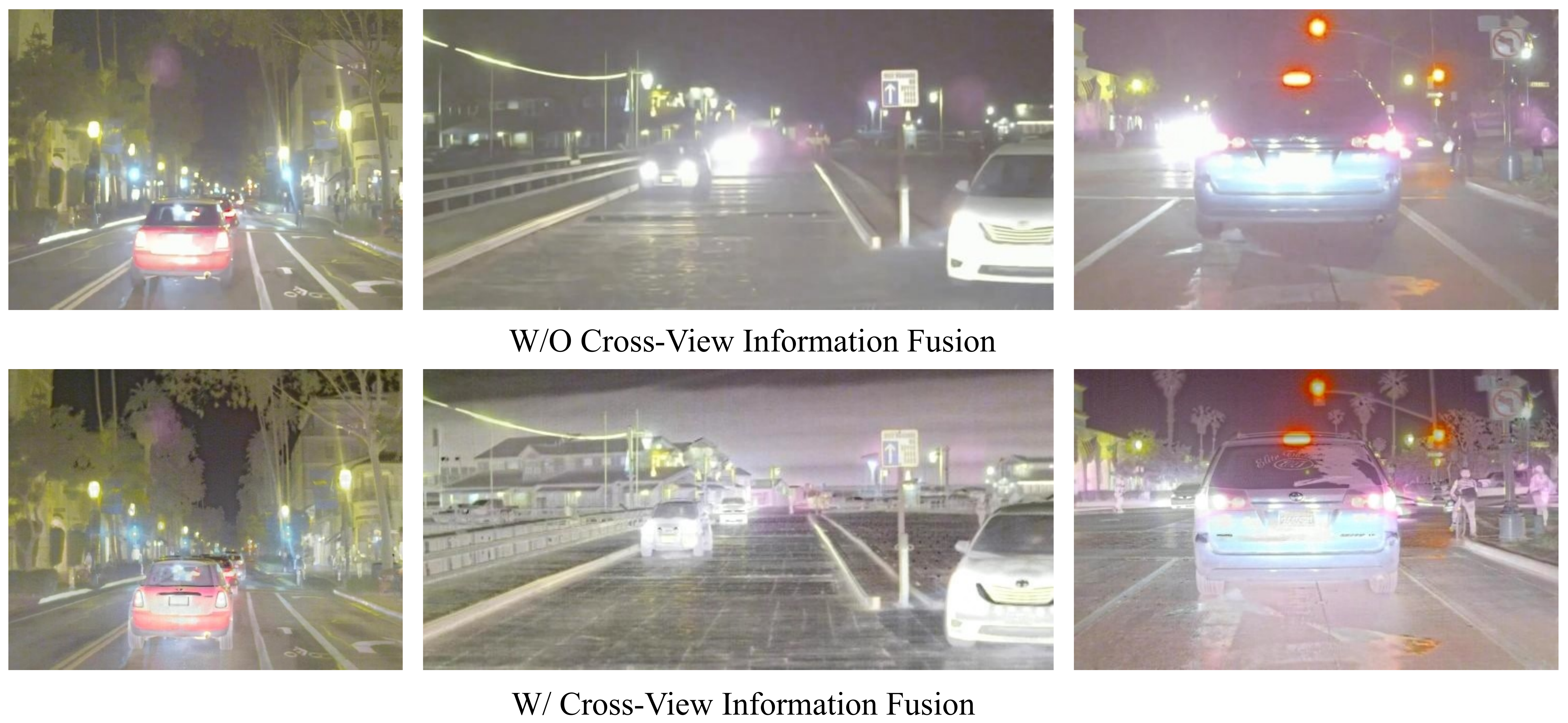}
\caption{Vision comparison of the ablation study on the second-stage FusionNet on three examples from our RoadScene-rain dataset.}
\label{fig10}
\vspace{-3mm}
\end{figure}

\begin{figure}[!t]
\centering
\includegraphics[width=0.95\columnwidth]{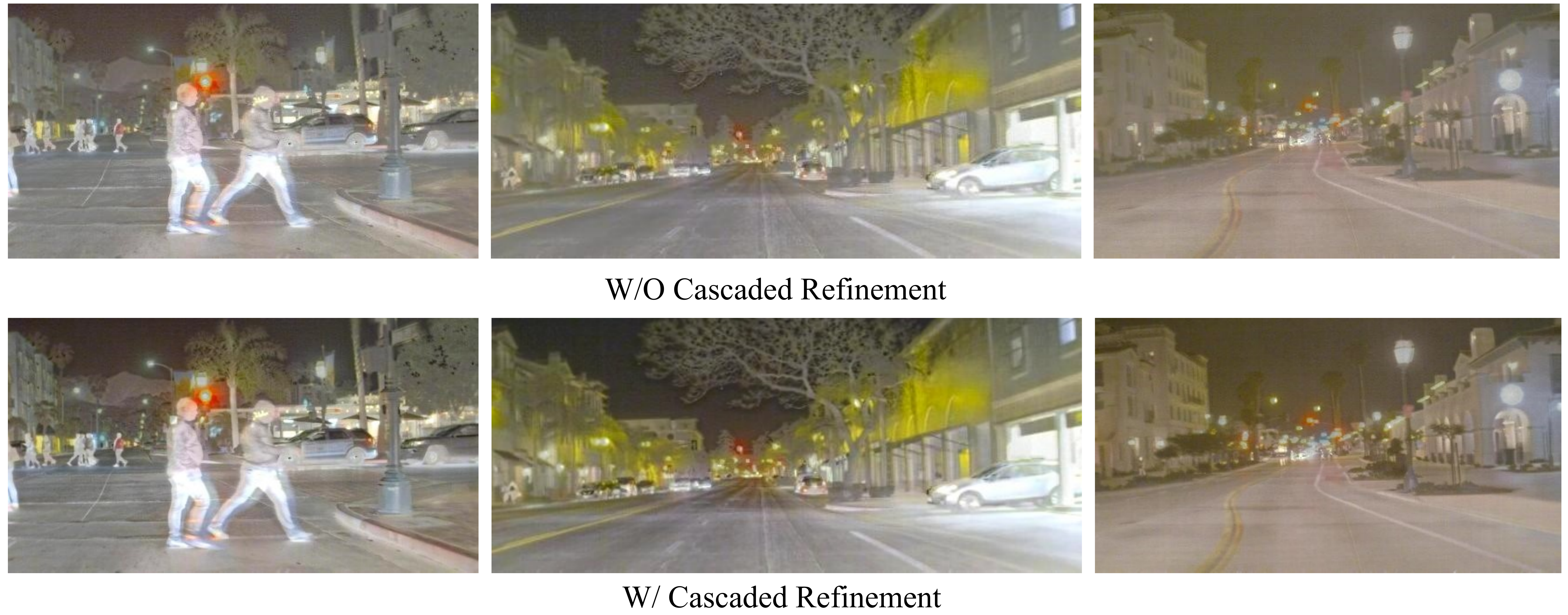}
\caption{Visualized results of ablation study on the cascaded refinement unit on three examples from RoadScene-rain dataset.}
\label{fig11}
\vspace{-3mm}
\end{figure}

\subsection{Ablation Studies}
In this part, we describe ablation studies to demonstrate the effectiveness of each component in our framework.

\subsubsection{Cross-view information fusion}
Cross-sensor information fusion enables our method to address the image de-raining task in low-light nighttime scenarios. As shown in Fig.~\ref{fig10}, it can be seen that without information fusion, de-rained images lack a comprehensive representation of driving scenes. Some crucial elements such as vehicles and pedestrians in low-light areas are challenging to capture. In contrast, our proposed approach effectively reconstructs high-quality results with clear structure, detailed textures, highlighted targets and enhanced contrast.  

\subsubsection{AgMoE}
We conduct an ablation study on the AgMoE module to evaluate its effectiveness, with results presented in Table~\ref{tab3}. Removing AgMoE from CleanNet results inleads to the underultilization of richer features for reconstructing clean-visible images, resulting in degraded fusion performance compared to our full model.

\subsubsection{TSSA}
The top-selection algorithm enables TSSA to better model long-range pixel dependencies for high-quality reconstruction. The corresponding ablation experiment, detailed in Table~\ref{tab3}, demonstrates the superiority of of our proposed method over the model without top-selection. This strategy empowers CleanNet to reconstruct finer detailed features, thereby contributing more effective and subtle texture details to the fusion process, ultimately improving the quality of fusion results.

\subsubsection{MDFFN}
We evaluate the effectiveness of MDFFN by comparing it with a conventional FFN. The results in Table~\ref{tab3} show that MDFFN significantly improves fusion performance by enriching textures in clean images through the extraction and fusion of local features at multiple scales.

\subsubsection{Information measurement}
The information measurement in the FusionNet is used to guide the fusion loss for enhanced fusion. We directly set $\omega_c$ and $\omega_{ir}$ to 0.5 in the fusion loss, rather than estimating them using information measurement values. The quantitative evaluation results are depicted in Table~\ref{tab3}. Obviously, our proposed method can preserve more source information and integrate richer complementary information, contributing to enhanced contrast and image quality in the fusion results.  

\subsubsection{Cascaded refinement}
 The cascaded refinement module facilitates the FusionNet to achieve contrast-improved and color-balanced fusion results. As illustrated in Fig.~\ref{fig11}, our proposed method with cascaded refinement can produce images providing improved contrast, brighter scenes, salient targets, and better visual effects. From the quantitative results shown in Table~\ref{tab3}, we can further conclude that our method reveals the concealed information and enhances the brightness and contrast in poor illumination.

\section{Conclusion}
We propose a novel cross-view sensor fusion network, which achieves de-raining in low-light nighttime scenarios through the joint component learning of cross-sensor data. Extensive experiments illustrate that our proposed framework has superior performance in both quantitative and qualitative results. Our method is well-suited for nighttime de-raining tasks, showcasing the effectiveness and improvement in rain removal, information complementation, and scene enhancement. To this end, our approach enhances the practical applicability of de-raining methods and advances their scope of generalization ability. For future work, we aim to gather diverse and complex real-world data to enhance the model performance in generalizing across various driving conditions.  Moreover, we will continue to improve the framework for real-time applicability while maintaining performance. We will also explore solutions to ensure robustness in tracking moving targets in dynamic environments. By pursuing these avenues, we aim to contribute to the development of more effective solutions for nighttime driving assistance systems.

\bibliographystyle{IEEEtran}
\bibliography{ref}

\end{document}